\documentclass[letterpaper]{article} 
\usepackage{aaai25}  
\usepackage{times}  
\usepackage{helvet}  
\usepackage{courier}  
\usepackage[hyphens]{url}  
\usepackage{graphicx} 
\usepackage{natbib}  
\usepackage{caption} 
\usepackage{algorithm}
\usepackage{algorithmic}
\usepackage{subcaption}
\usepackage{microtype}
\usepackage{multirow}
\usepackage{float}
\usepackage[figuresright]{rotating}
\usepackage{graphicx}		
\usepackage{makecell}
\usepackage{threeparttable}
\usepackage{rotating} 

\usepackage{listings}
\usepackage{xcolor}
\usepackage[utf8]{inputenc}
\definecolor{darkblue}{RGB}{0, 0, 139}

\usepackage{newfloat}
\usepackage{listings}
\DeclareCaptionStyle{ruled}{labelfont=normalfont,labelsep=colon,strut=off} 
\floatstyle{ruled}
\newfloat{listing}{tb}{lst}{}
\floatname{listing}{Listing}
\usepackage{amsmath}
\usepackage{threeparttable}
\usepackage{xcolor}

\title{Consistency of Responses and Continuations Generated by Large Language Models on Social Media}

\author{
    Wentao Xu\textsuperscript{\rm 1}\thanks{Corresponding author, myrainbowandsky@gmail.com}\equalcontrib,
    Wenlu Fan\textsuperscript{\rm 2}\equalcontrib,
    Yuqi Zhu\textsuperscript{\rm 3},
    Bin Wang\textsuperscript{\rm 4},    
}
\affiliations{
    \textsuperscript{\rm 1,2} Department of Science and Technology of Communication,\\ \textsuperscript{\rm 3} School of Humanities and Social Sciences, \\  \textsuperscript{\rm 4} Independent Researcher, Beijing, 100000, China
   \textsuperscript{\rm 1,2,3} Univesity of Science and Technology of China\\
    No.96, JinZhai Street, Hefei, Anhui, 230026, China\\
}

\begin{document}

\maketitle

\begin{abstract}
Large Language Models (LLMs) demonstrate remarkable capabilities in text generation, yet their emotional consistency and semantic coherence in social media contexts remain insufficiently understood. 
This study investigates how LLMs handle emotional content and maintain semantic relationships through continuation and response tasks using three open-source models: Gemma, Llama3 and Llama3.3 and one commercial Model:Claude. By analyzing climate change discussions from Twitter and Reddit, we examine emotional transitions, intensity patterns, and semantic consistency between human-authored and LLM-generated content. Our findings reveal that while both models maintain high semantic coherence, they exhibit distinct emotional patterns: these models show a strong tendency to moderate negative emotions. When the input text carries negative emotions such as anger, disgust, fear, or sadness, LLM tends to generate content with more neutral emotions, or even convert them into positive emotions such as joy or surprise. At the same time, we compared the LLM-generated content with human-authored content. The four models systematically generated responses with reduced emotional intensity and showed a preference for neutral rational emotions in the response task. In addition, these models all maintained a high semantic similarity with the original text, although their performance in the continuation task and the response task was different. These findings provide deep insights into the emotion and semantic processing capabilities of LLM, which are of great significance for its deployment in social media environments and human-computer interaction design.
\end{abstract}

%

\section{Introduction}
Large Language Models (LLMs) represent one of the most significant yet controversial technological advancements in recent years. These models demonstrate unprecedented and expanding human-like capabilities, particularly in text generation, enabling diverse applications including text summarization \cite{van2024field}, translation \cite{sung2024context}, and news writing \cite{munoz2024contrasting}. Consequently, LLM-based applications have proliferated across domains, from conversational agents \cite{dam2024complete} to educational assistants \cite{liu2025llms}.

Despite their advantages, LLMs raise significant concerns regarding potential negative implications. These include content fabrication, commonly termed ``hallucination,'' which contributes to misinformation propagation \cite{huang2023survey}. Furthermore, research indicates that LLM-generated content perpetuates societal biases encountered during training, potentially exacerbating AI fairness issues \cite{gallegos2024bias, ayoub2024inherent}. Additionally, LLMs can influence human decision-making processes, potentially leading to unintended consequences through emotional manipulation or deception \cite{park2024ai}. Given their widespread deployment, careful evaluation of LLMs' text generation capabilities becomes imperative.

LLMs exhibit both task-specificity and context-sensitivity, with performance varying across different applications and contextual settings \cite{sung2024context, li2023metaagents}. Consequently, evaluating their text generation capabilities within realistic, socially relevant contexts becomes crucial. Social media platforms, serving as extensive networks for information exchange, provide valuable digital artifacts for such investigations.


In social media contexts, LLM text generation manifests in two primary forms: response tasks (e.g., replies) and continuation tasks (e.g., summarization and dialogue). The generated content influences public perception and engagement on social media platforms. Emotion embedded within text plays a crucial role as it can be rapidly activated and disseminated through extensive social networks, potentially facilitating emotional contagion \cite{kramer2014experimental}. Consequently, emotion serves as a strategic tool for engagement and persuasion in social media environments \cite{stieglitz2013emotions, hamby2022effect}.

Previous investigations of emotional effects on social media have employed real-life experiments through content manipulation \cite{kramer2014experimental}. However, it raise ethical concerns regarding manipulation of user content and public discomfort \cite{boyd2016untangling}.  Recent work demonstrates that LLM-based simulations can be a useful tool for modeling social interactions \cite{gao2024large}: (1) AI agents can serve as safer substitutes for human participants in extreme or sensitive scenarios, and (2) they enable more controlled experimental conditions, facilitating precise examination of relevant variables.

With the widespread adoption of LLMs in generating human-like content, it becomes imperative to understand the consistency of LLM-generated text and its potential societal impact. Moreover, with a growing variety of LLMs being adopted in real-world settings, there remains a lack of systematic, comparative evaluations of how consistent/different these models perform.

Accordingly, this study investigates LLM text generation tasks (response generation and continuation) through systematic analysis of emotional consistency and semantic similarity. By examining these dynamics within climate change communication—a highly polarized and emotionally charged domain—this research addresses the following questions:

\textbf{\textit{RQ1: How consistent are the emotions expressed in text generated by LLMs on social media? }}

\textbf{\textit{RQ2: How does the emotional intensity of text generated by LLMs compare to text on social media? }}

\textbf{\textit{RQ3: To what extent do LLMs demonstrate semantic similarity between generated text and text on social media?}}

By answering the research questions, this study has the following findings and contributions:
\begin{itemize}
\item We systematically evaluated emotional and semantic consistency between LLM-generated and human-authored social media texts, finding that (1) LLMs’ emotional expressions differ significantly from original texts across both continuation and response tasks, and across different model types; (2) LLMs maintain high semantic similarity with original content, indicating strong capabilities in understanding and generating human-like text.

\item We found that all models exhibit significantly lower emotional intensity in both tasks, suggesting that LLMs may struggle to convey the full emotional depth of human-authored content.

\item We discussed the real-world implications of these emotional differences and semantic proximity — including how LLMs may influence emotional engagement, serve as tools for moderating polarized debates, or, conversely, be misused for emotional manipulation.

\item We proposed an ethically grounded simulation framework using LLM agents to explore emotional dynamics around controversial topics on social media, avoiding privacy and consent issues inherent in traditional research \cite{ferrara2015measuring}. We also adopted the “LLM-as-judge” approach to assess semantic similarity, reducing reliance on time-consuming and costly human annotations.
\end{itemize}

\section{Related Works}
\subsection{Evaluation of LLMs generated text}
The evaluation of LLM-generated text originates from natural language generation (NLG), defined as the process of computationally producing human-comprehensible text \cite{sai2022survey}. Given the widespread deployment of AI models in text generation, extensive research has explored effective evaluation frameworks for NLG \cite{sai2022survey}.
Traditional evaluation metrics, primarily focused on quantifying content overlap between system outputs and references \cite{gao2024llm}, such as BLEU \cite{papineni2002bleu} and ROUGE \cite{lin2004rouge}, have served as standard metrics for automatically assessing output quality in machine translation and summarization tasks. However, these metrics demonstrate limitations when applied to complex, context-dependent tasks, particularly in the current generative AI paradigm \cite{gao2024llm}. Consequently, researchers have developed novel benchmarks for task-specific LLM evaluation (e.g., \cite{que2024hellobench}), while recent studies have proposed methodologies leveraging LLMs themselves for evaluation purposes (see \cite{gao2024llm} for a comprehensive review).

The evaluation of LLM-generated text consistency with human behavior represents a fundamental approach to assessing model performance. Alignment with human behavior and response patterns remains a central objective in artificial intelligence development \cite{russell2016artificial}. Consistency is crucial for operational reliability and safety of LLMs, ensuring they can generate contextually appropriate and relatable outputs. Additionally, semantic similarity serves as an established metric for quantifying textual consistency \cite{chandrasekaran2021evolution}. Researchers have evaluated LLM output consistency through semantic similarity measures and developed enhancement strategies to improve human alignment \cite{yang2024enhancing, raj2023semantic}.

Existing literature predominantly examines distinctive characteristics between LLM- and human-generated text. For instance, \cite{herbold2023large} conducted comparative analyses of human-written versus ChatGPT-generated essays across dimensions including topical coverage, logical structure, vocabulary usage, and linguistic constructions through human assessment. Beyond manual annotation, \cite{guo2023close} implemented a mixed-methods approach to analyze LLM/human-generated responses across linguistic dimensions, revealing that LLM outputs demonstrate enhanced logical coherence, comprehensive detail, and reduced bias. \cite{munoz2024contrasting} employed quantitative analysis to compare human- and LLM-authored news content across morphological, syntactic, psychometric, and sociolinguistic dimensions. Through automated analysis, \cite{zanotto2024human} identified distinctive linguistic patterns in text length, variability, syntactic complexity, and lexical diversity.

\subsection{Text generation on social media context}
In the social media environment, LLM text generation offers significant applications, including AI-powered social bots for online discourse participation, discussion summarization tools, and related applications \cite{li2024pre}. However, ensuring generated text consistency requires careful consideration of contextual factors and interaction objectives. Social media interactions encompass both response generation (e.g., comment replies) and content continuation (e.g., social bot engagement). While existing research provides empirical evidence comparing human and LLM-generated content, the evaluation of social media-specific tasks, particularly responses and continuations, warrants comprehensive evaluation to understand LLM text generation in dynamic social media contexts.

Although emotion serves as a crucial factor in social media engagement and persuasion, its utilization as an evaluative feature for text generation remains insufficiently explored. Current comparative studies of human and LLM-generated text focus predominantly on static contexts, overlooking emotional dynamics. For instance, comparative analysis of human-written versus LLM-generated news content revealed stronger negative emotional expression in human-authored texts \cite{munoz2024contrasting}. Similarly, while \cite{guo2023close} examined response differences through multilingual sentiment classification, this approach presents limitations for comprehensive emotional analysis (e.g., distinguishing between joy and sadness). Given the dynamic nature of social media interactions, evaluation of emotional consistency in information exchange becomes crucial.
Emotional content permeates social media discourse and functions as a crucial determinant in shaping public opinion \cite{naskar2020emotion}.
Emotion demonstrates high susceptibility to influence and serves as a critical factor in controversial and uncertain social agendas, including epidemics \cite{lu2022emotional}, disasters \cite{chu2024emotional}, and polarizing social issues such as climate change \cite{brady2017emotion}. In these contexts, emotional responses can exert both beneficial and detrimental effects on public discourse. Climate change discourse, in particular, represents an extensively studied yet remains highly polarized domain, characterized by persistent denialism and skepticism \cite{treen2020online, whitmarsh2011scepticism}. These misconceptions frequently leverage emotional appeals, particularly fear, to influence public perception \cite{martel2020reliance}. Clinical psychology research has established correlations between anger, elevated cortisol responses to stress, and increased vulnerability to misinformation \cite{sharma2023systematic}.

Above all, evaluating emotional patterns across response and continuation tasks within climate change discussions on social media provides a crucial framework for comparing LLM and human-generated content in dynamic, real-world scenarios.


\section{Methodology}

\subsection{Experimental Design}
\begin{figure*}[htb]
\centering 
\includegraphics[width=0.6\linewidth, height=0.3\textwidth]{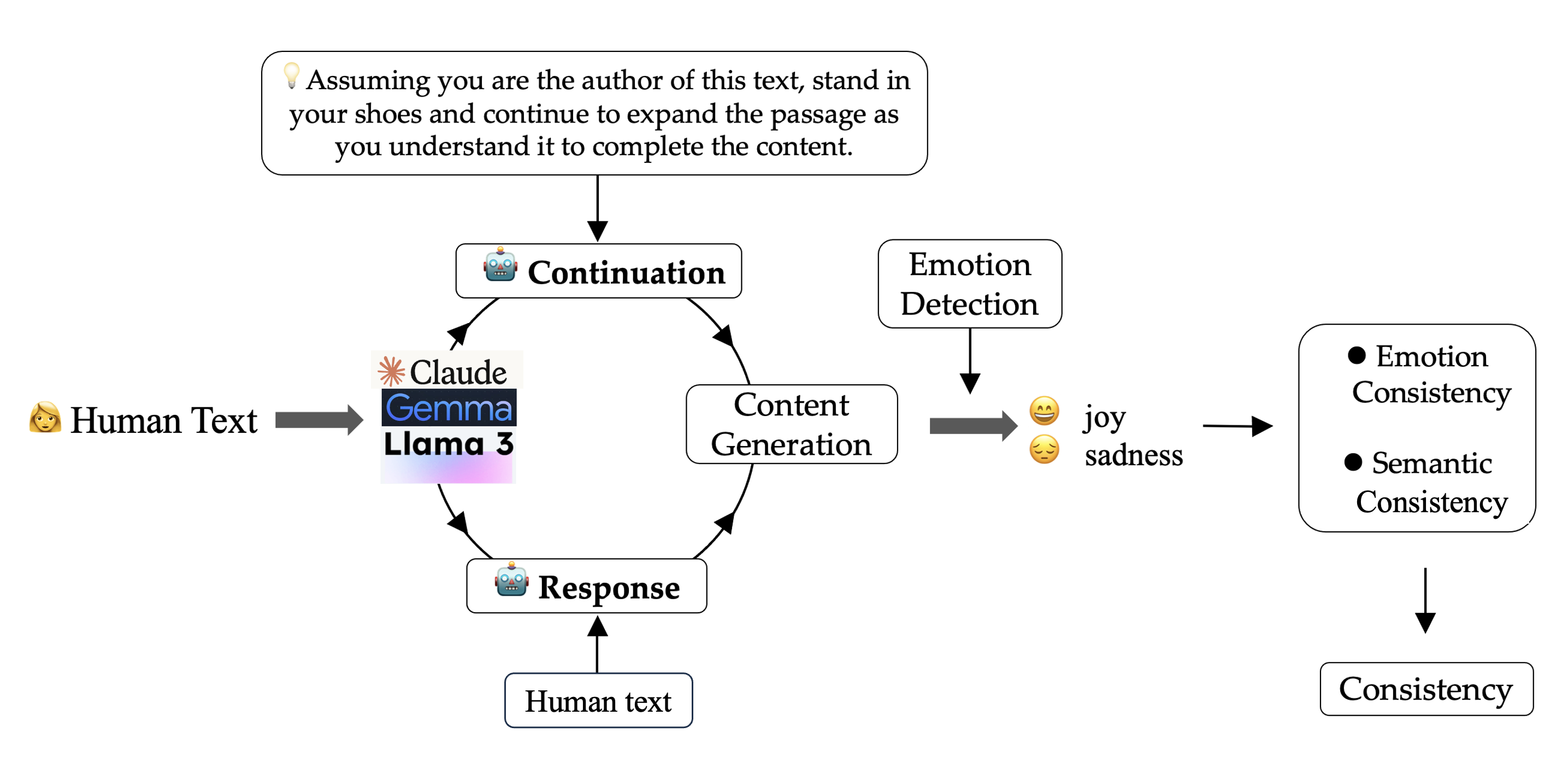}
\caption{Experimental pipeline of consistency evaluation for LLMs. Our experimental framework begins with human text input to four LLMs , which perform two distinct tasks: continuation and response. The continuation task employs a specific prompt instructing the model to expand the text as its author, while the response task operates without explicit prompting to enable natural interaction. Following content generation, we implement emotion detection on the outputs, followed by comprehensive analyses. The framework concludes with parallel analyses of emotional content and semantic consistency to evaluate the consistency of LLM-generated content relative to the original human input.}
\label{Experiment design}
\end{figure*}

Figure~\ref{Experiment design} illustrates the overall setup of the experimental design. 
Our core experiment involves using LLMs to interact with human-generated text data, which consists of social media posts collected from real users.  
In our experiment, we employed three open-source large language models—Gemma\footnote{https://ai.google.dev/gemma}, Llama 3\footnote{https://ai.meta.com/llama/license/}, and Llama 3.3—developed by Google and Meta, respectively, as well as one commercial model, Claude 3.5, developed by Anthropic\footnote{https://www.anthropic.com/claude}. Specifically, we utilized the following model variants: Gemma2-27B-Instruct-Q8, Llama3-70B-Instruct, Llama3.3-70B, and Claude-3-5-Haiku-20241022-X. These models were selected due to their strong performance and robustness. We utilized Ollama\footnote{https://ollama.com/} as a framework to enable the two open source models to run on our local server.

In our study, we mainly used the chat of Ollama.
For the response task, we directly used the chat interface, which allows LLMs to interact with the input content without requiring explicitly crafted instruction prompts. This setup offers a more natural and direct observation of the LLM’s behavior as a conversational agent. However, we acknowledge the reviewer’s point and clarify that although we did not provide any additional instruction prompt, the original human post itself serves as the input prompt to the model. For the continuation task, we used the content generation function, where concise prompts were employed to minimize their influence on the model, allowing for a clearer focus on the model's response.We tell the LLMs that \textit{``Assuming you are the author of this text, stand in your shoes and continue to expand the passage as you understand it.''}

\subsection{Dataset}
This study utilized climate change corpora collected from Twitter (now X) and Reddit. We collected data using the Twitter Search API by querying relevant keywords, including ``climate change'', ``climate science'', ``climate manipulation'', ``climate Engineering'', ``climate Hacking'', ``climate modification'', ``Global Warming'', ``carbon footprint'', and ``The Paris Agreement''. For Reddit, we used data maintained by Pushshift from https://the-eye.eu/redarcs/. The Pushshift Reddit dataset consists of two sets of files: submissions and comments ~\cite{DBLP:journals/corr/abs-2001-08435}. The same keywords were applied to filter Reddit data, and to compare the differences in emotions, we collected both posts and comments from both platforms.

\begin{figure*}[ht]
\begin{minipage}[ht]{0.5\linewidth}
\centering
\includegraphics[width=0.7\textwidth, height=0.5\textwidth]{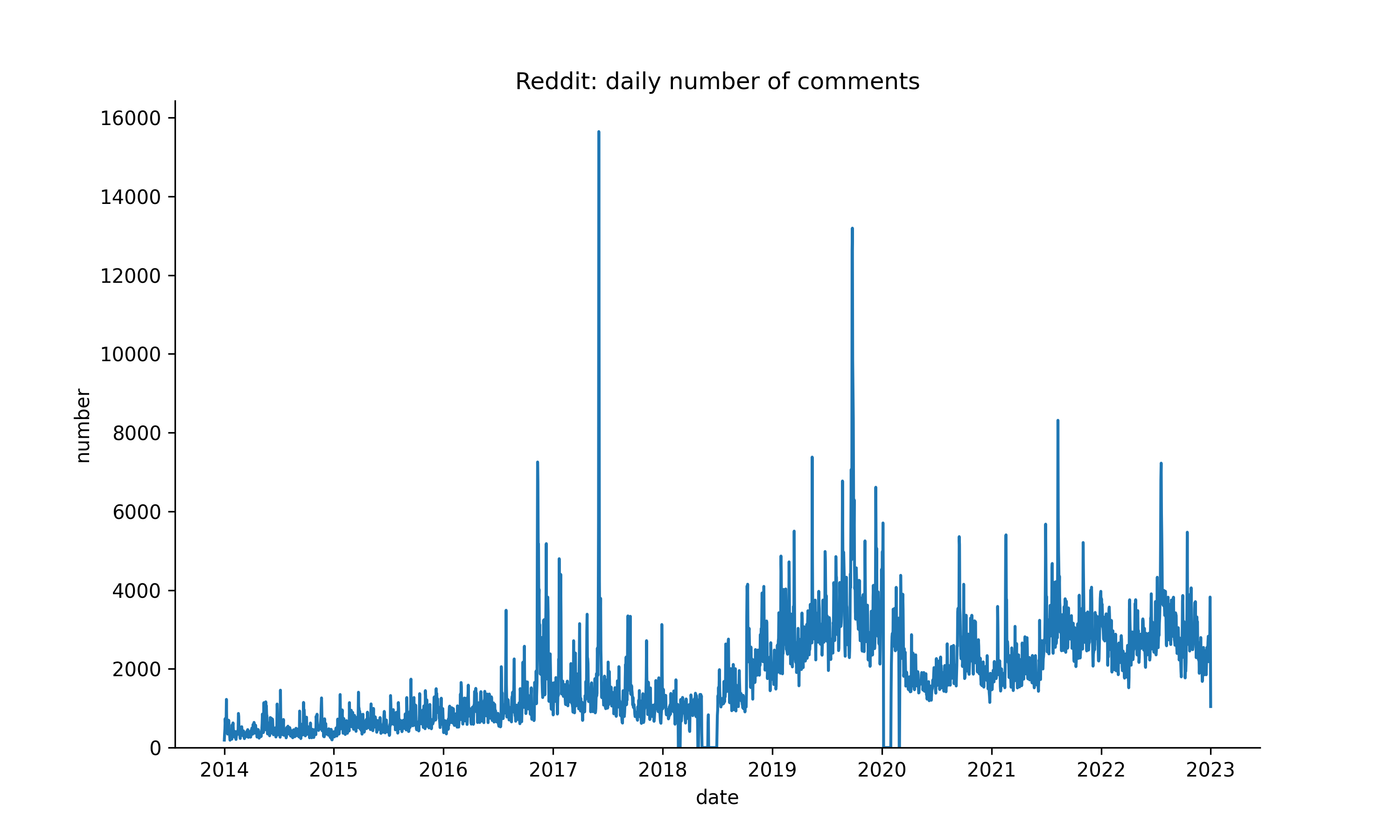}
\centerline{\textbf{a}}
\end{minipage}%
\begin{minipage}[ht]{0.5\linewidth}
\centering
\includegraphics[width=0.7\textwidth, height=0.5\textwidth]{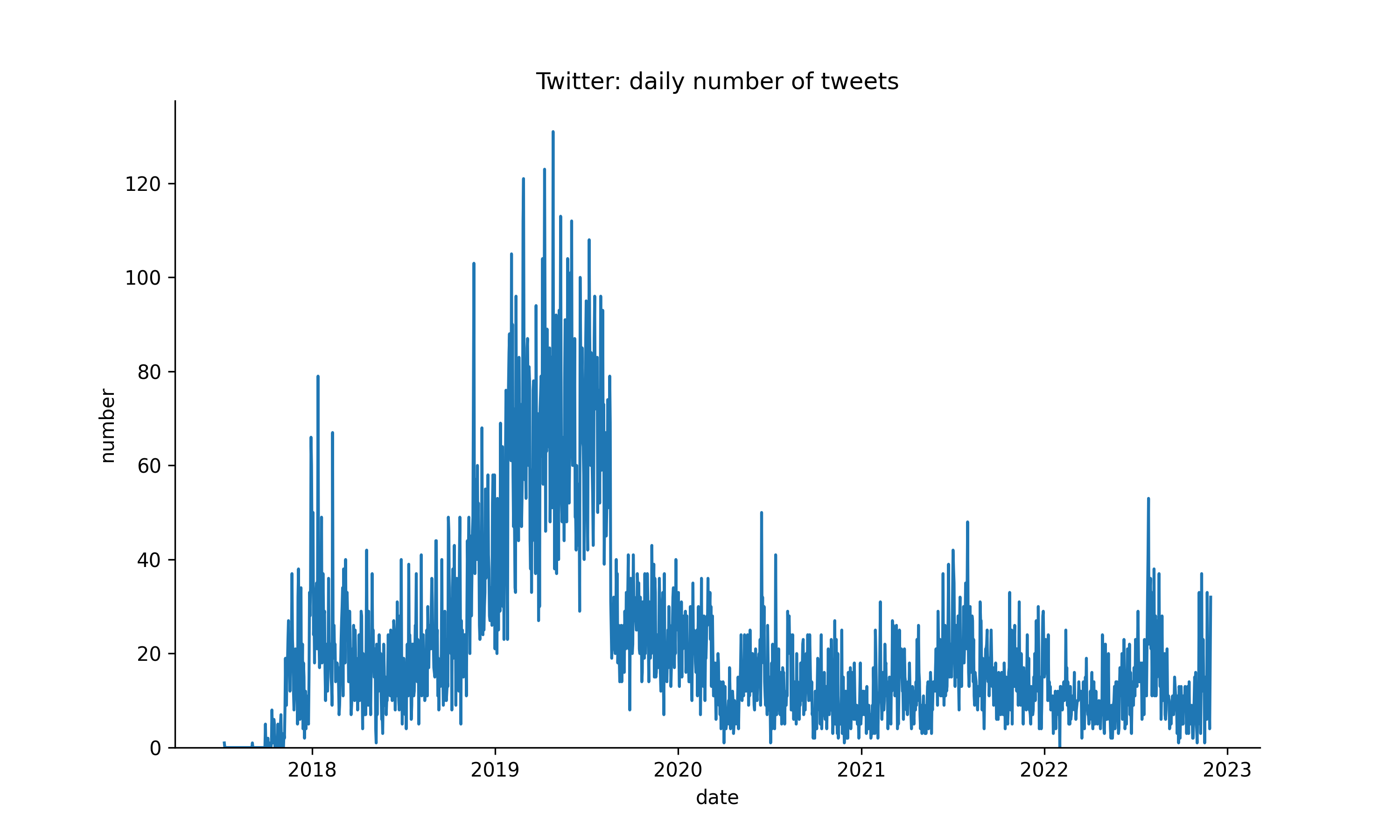}
\centerline{\textbf{b}}
\end{minipage}
\caption{Daily data amount of Twitter and Reddit.
\textbf{a.} Daily comments count of Reddit. \textbf{b.} Daily tweets count of Twitter. The x-axis represents the date, and the y-axis represents the frequency.}
\label{daily distribution}
\end{figure*}

With the keywords, we obtained 5,768,822 Reddit comments and 76,596,654 tweets from Twitter. We used histograms to understand the basic distribution of data (Figure~\ref{daily distribution}). 
Twitter and Reddit are two major social media platforms that differ from each other. Twitter is designed to be open, concise, and immediate through short posts and real-time updates, while Reddit is a community-based medium designed for deep, deliberative, and topical discussions\cite{10216752, Treen04072022}. These differences in platform design not only shape how content is generated and shared, but also attract different user groups with different interests and engagement styles, which provides a valuable resource for understanding different modes of public discussion\cite{ruan2022cross}.

We support these distinctions with some quantitative evidence using the average word length, word cloud. The results show that the average length of Reddit text is 656.66 words, while the average length of Twitter text is 246.34 words. This discrepancy reflects fundamental differences in how the two platforms are used. Reddit is structured around topic-based discussions within subreddits, encouraging longer, more detailed posts that often resemble mini-essays or narratives. The word cloud chart (see the appendix\ref{wordcloud}) shows that the expressions of users on the two platforms under the same topic are also different. For example, more people on Reddit are discussing "climate change", while more people on Twitter use "global warming".

To construct datasets representative of the discussion dynamics surrounding climate change over an extended period and to mitigate potential biases introduced by sudden climate change events, we employed a time-stratified sampling approach on the raw data. Specifically, based on the collected corpus of 5,768,822 Reddit comments and 76,596,654 Twitter tweets, we initiated the process by partitioning the data by month. Subsequently, within each monthly segment, we conducted systematic sampling, extracting 200 records from the Twitter data and 100 records from the Reddit data. This method of monthly proportional sampling was implemented to ensure a uniform distribution of the dataset across the temporal dimension, thereby reducing the influence of transient extreme events or fluctuations in topic popularity during specific timeframes. This strategy aims to enable the final dataset to more comprehensively reflect the content and evolving trends of discussions pertaining to climate change across different temporal stages. Ultimately, this process yielded an analytical dataset comprising 12,200 Twitter data points and 10,900 Reddit data points.

\subsection{Emotion Labeling}
In this study, we developed a methodology to analyze emotions in cross-platform social media data using a deep neural network-based model. We employed the\textit{j-hartmann/emotion-english-distilroberta-base\footnote{https://huggingface.co/j-hartmann/emotion-english-distilroberta-base}} model from Hugging Face to examine the emotional content of both original texts and content generated by large language models. This model, built upon the RoBERTa-base architecture, is a fine-tuned checkpoint of ``DistilRoBERTa-base" the datasets contain emotion labels for texts from Twitter, Reddit, student self-reports, and utterances from TV dialogues.

The \textit{emotion-english-distilroberta-base} model identifies seven distinct emotion categories:joy, surprise, neutral, anger, disgust, fear and sadness. The model outputs probability scores for each category, which serve as quantitative measures of emotional content for subsequent analysis. While previous studies used sentiment analysis (negative, positive, and neutral) for evaluating differences between human and LLMs-generated text \cite{guo2023close}, our emotion-based approach provides a more granular and nuanced understanding of the underlying emotional states in posts.



\subsection{Semantic Consistency}

We design a evaluation framework that uses a rubric-guided expert-like LLM to score consistency and faithfulness. Here we hired the model:gemini-2.0-flash-thinking-exp-1219 as our expert LLM. This allows us to directly assess whether a model-generated response is logically aligned with the user's comment, and please see the appendix for specific prompt design\ref{lst:build-prompt}.
After obtaining the evaluation results of the LLM-as-judge, the results are reviewed by a human evaluator. Human evaluator judges whether the evaluation results of the LLM are reasonable and accurate. In the end, the consistency rate between the two is maintained at 75\%.

Automatic evaluations of instruction following abilities in LLMs has recently received significant attention~\cite{zheng2023judging}. Given the significant time and cognitive effort required for human evaluation of large-scale generated content, we adopted a stratified random sampling approach to ensure representativeness across different model-platform-task combinations. Our full dataset consists of different sub-corpora, each corresponding to a specific combination of model (e.g., Gemma, Llama 3, Llama 3.3, Claude), platform (Twitter, Reddit), and task type (response or continuation), with approximately 10,000 instances per sub-corpus.
To balance coverage and feasibility, we randomly sampled 30 instances from each sub-corpus, resulting in a total of 480 samples for evaluation.

Manual expert evaluation is costly. This hybrid evaluation strategy combines manual judgment with the assistance of a LLM. Although the sample size is small, it covers diversity across platforms, tasks, and models. We acknowledge in the limitations section that this small sample size may affect the generalizability of our findings.

\section{Results}
\subsection{Emotion Dynamics of the Original Text in Downstream Tasks}

In this study, we examined the emotional transitions between human-generated text and LLM outputs in downstream tasks. We categorize 7 emotions as positive emotionally oriented and negatively oriented as well as neutral emotions as defined by itself, as follows~\cite{robinson2008brain} :

Positive emotions: anticipation, joy, love, optimism, surprise, trust ~\cite{vaillant2008positive};

Negative emotions: anger, disgust, fear, pessimism, sadness

\begin{figure*}[ht]
\centering
\includegraphics[width=1\linewidth]{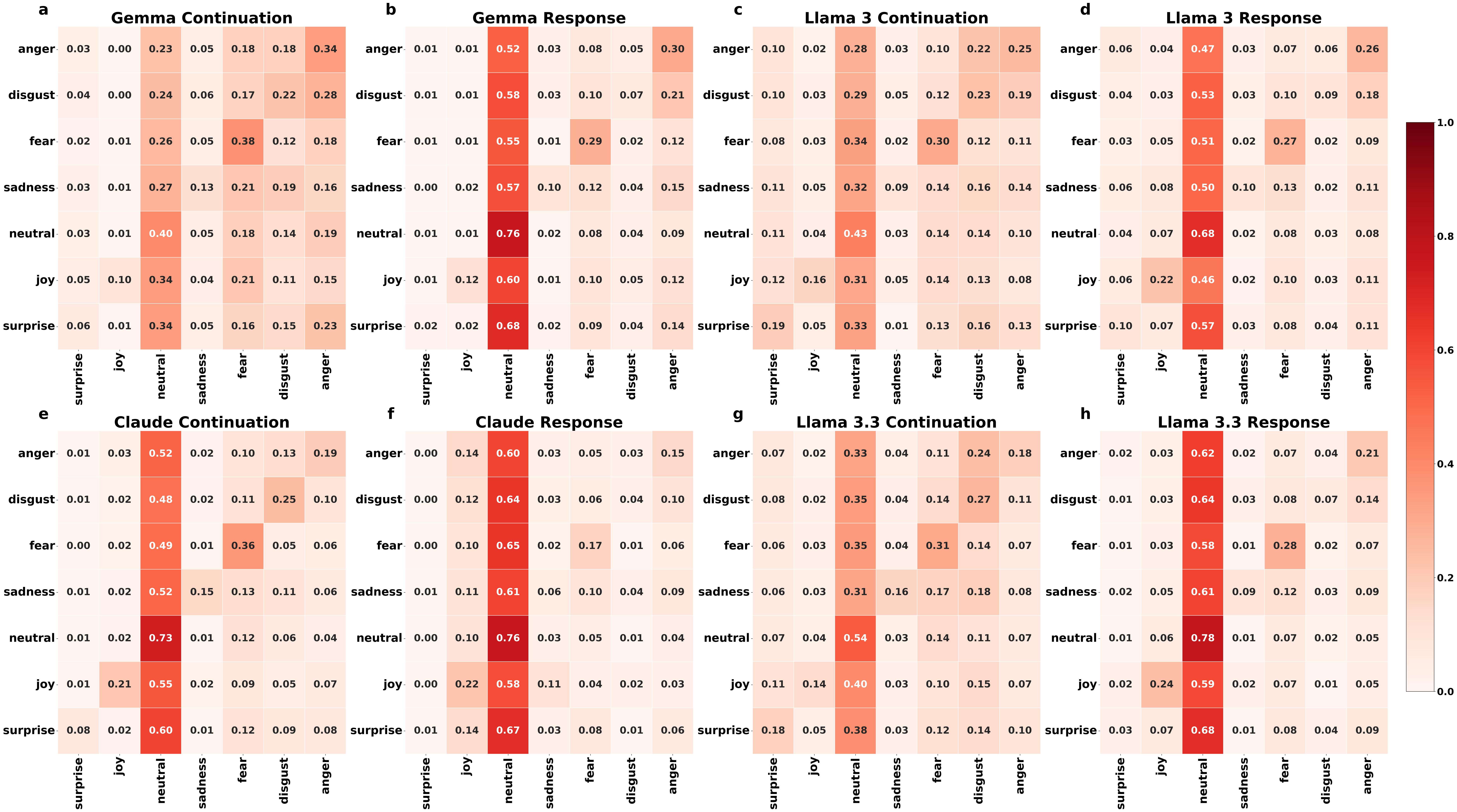}
\caption{Emotional Transition Analysis of LLM Response and Continuation Tasks in Reddit Comments.
Panels \textbf{a}, \textbf{b}, \textbf{c}, \textbf{d}, \textbf{e}, \textbf{f}, \textbf{g}, and \textbf{h} illustrate emotional transitions in content generated by Gemma, Llama and Claude models during continuation and response tasks, respectively. The y-axis represents source emotions from human text, while the x-axis indicates emotions in LLM-generated content. Cell values represent the proportion of emotional transitions between original and generated content. For example, in Figure \textbf{3a}, the value 0.34 in the anger-to-anger cell indicates that 34\% of originally angry texts maintained their emotional valence in Gemma's continuation task. The intensity of each cell's shading represents the proportion of emotional transition, with darker shades indicating higher transition frequencies.}
\label{Reddit:heatmap}
\end{figure*}
Analysis of Figure 3a reveals that in Gemma's continuation tasks, 34\% of the texts initially labeled as 'anger' maintained their emotional, and 38\% of the texts originally categorized as 'fear' also preserved the 'fear' emotion in the continuation task. For other emotion categories, including 'disgust,' 'sadness,' 'neutral,' 'joy,' and 'surprise,' the proportions of texts that retained the same emotional label were 22\%, 13\%, 40\%, 10\%, and 1\%, respectively, demonstrating a capability for emotion preservation.

It is noteworthy that for all original emotion labels, a considerable proportion of the texts shifted to a neutral label in the continuation task. Concurrently, a significant portion of texts also transitioned to the 'anger' label during the continuation. These findings suggest that the Gemma model possesses a certain ability to recognize and perpetuate the original emotions. In subsequent tasks, it exhibits a systematic tendency to convert various emotional expressions into a neutral sentiment. The results in Figure 3a also, to some extent, indicate the model's sensitivity to negative emotions in the continuation task, particularly anger.

In analyzing the emotional shifts within Gemma's response tasks, we observed significant changes in emotional conversion. For texts with originally positive emotions such as 'joy' and 'surprise,' as well as those with negative emotions like 'anger,' 'disgust,' 'fear,' and 'sadness,' over 50\% of their emotion labels were converted to neutral in the response task. Our analysis reveals a systematic bias in Gemma towards rational, neutral sentiment in these response tasks. Notably, a portion of the original emotions still transformed into 'anger' in the responses; for instance, 'disgust' showed a 21\% conversion rate to 'anger,' while 'surprise' had a 14\% conversion rate. This suggests Gemma's sustained sensitivity to anger-related content across both response and continuation tasks.

Figure~\ref{Reddit:heatmap}c and Figure~\ref{Reddit:heatmap}d illustrate the performance of Llama 3, showing that it has the same ability to recognize and preserve emotions in continuation and response tasks, and is able to maintain the original emotional valence more consistently. Similarly, we see a neutral emotion bias in the Llama3 model.

In the response and continuation tasks of the commercial model Claude, we observed that more original text emotions were more neutral in downstream tasks. At the same time, in the same model family, we observed that the Llama 3.3 model was also more inclined to generate neutral content in response and continuation than Llama3.

\begin{figure*}[ht]
\centering
\includegraphics[width=1\linewidth]{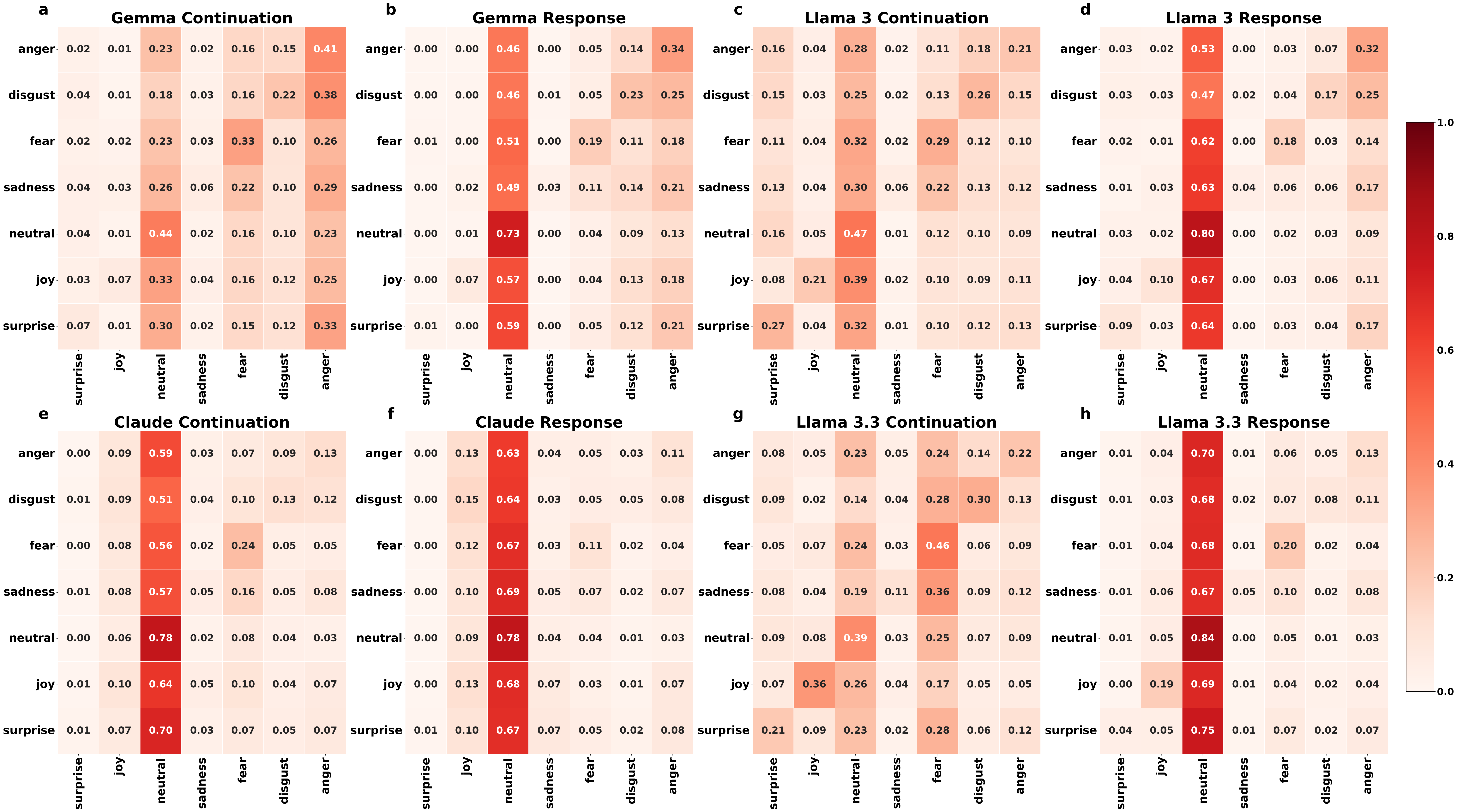} 
\caption{Emotional Transition Analysis of LLM Response and Continuation Tasks in Twitter Comments. Panels \textbf{a}, \textbf{b}, \textbf{c}, \textbf{d} \textbf{e}, \textbf{f}, \textbf{g}, and \textbf{h} illustrate emotional transitions in content generated by Gemma, Llama and Claude models during continuation and response tasks on Twitter, respectively. The y-axis represents the original emotions in human-authored tweets, while the x-axis shows the emotions detected in LLM-generated content. Each cell value represents the proportion of emotional transitions, with darker shades of red indicating higher transition frequencies. }
\label{Twitter:heatmap}
\end{figure*}

Beyond Reddit, our analysis also included Twitter data, which revealed distinct discourse patterns surrounding climate change topic. Figure~\ref{Twitter:heatmap} illustrates the emotion shift between the original Twitter content and the replies generated by the LLM.
The emotion performance patterns of the Gemma, Llama and Claude models on Twitter texts in continuation and response tasks yielded two key insights: Firstly, Gemma demonstrated heightened sensitivity to content related to anger; secondly, for models exhibited a systematic bias towards neutral sentiment when operating in interactive conversational contexts. third, both Claude and Llama3.3 models tend to generate content with neutral emotions. In addition, it is worth noting that in the continuation task of Llama3.3, its ability to continue emotions is better than Llama3.

\subsection{Resources of LLMs' Generated Content Emotions}

\begin{figure}[htbp]
\centering
\includegraphics[width=\linewidth]{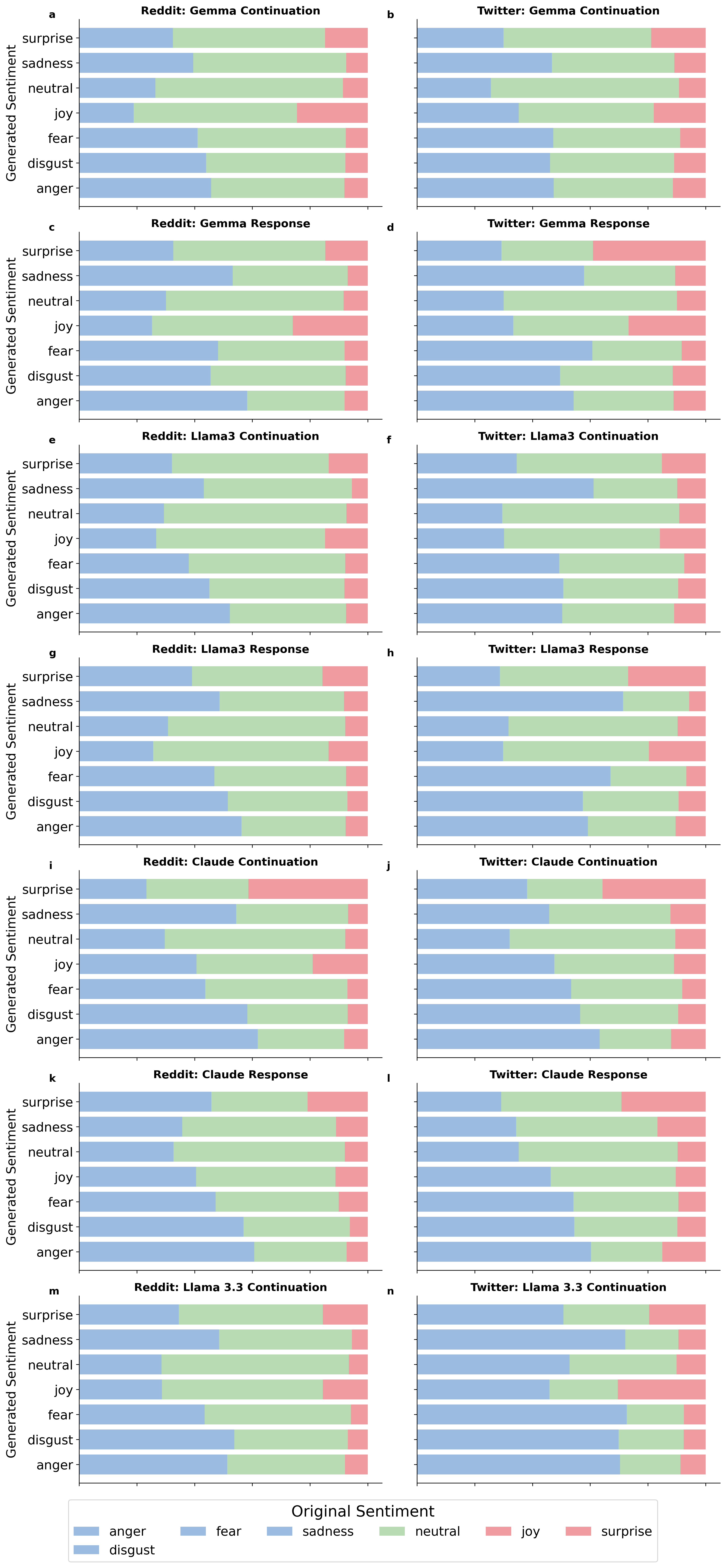} 
\caption{Emotional source analysis of LLM-generated content across platforms. Panels \textbf{a}--\textbf{n} illustrate emotional transitions in Gemma, Llama and Claude models' continuation and response tasks on Reddit and Twitter data. The red bars indicate that the original text corresponding to the generated text is positive emotions such as joy and surprise, the blue bars indicate that the original text is negative emotions such as anger, disgust, fear and sadness, and the green bars indicate that it comes from text with neutral emotions. The y-axis displays the emotional categories present in both original and generated content.}
\label{Reddit:emotion sources}
\end{figure}

We analyzed the emotional sources of LLM-generated content by examining the relationship between input and output emotions.
As shown in Figure~\ref{Reddit:emotion sources}a, the results show that Gemma did not simply copy or continue the original emotion of the input text, but perform significant emotion modulation. A core finding is that the model has a general tendency to moderate negative initial emotion. Specifically, when the original text carries negative emotion (as shown by the blue bar in the figure~\ref{Reddit:emotion sources}, representing anger, disgust, fear, or sadness), a considerable proportion of the generated content emotion will turn into neutral (source shown by the green bar) or even positive emotion (source shown by the red bar, representing joy or surprise). For example, in the continuation and response tasks of multiple models (such as Gemma, Llama, Claude), as well as on the Reddit and Twitter datasets, it can be observed that the generated text with "neutral" or "joy" emotion has a significant negative emotion contribution in its original emotion input. In addition, the model also shows a general trend towards neutral emotion, that is, the generated neutral emotion text comes from a wide range of original positive, negative, and neutral inputs. Even if the original input is positive, the generated content has a certain probability of being adjusted to neutral. These emotion conversion patterns show high consistency between different LLMs, task types (text continuation and response), and data sources (Reddit and Twitter), suggesting that this may be an inherent characteristic or common training result of current mainstream LLMs in emotion processing.

\subsection{Comparative Analysis of Emotional Intensity between LLMs and Human Text}
We analyzed the differences in emotional intensity between LLM-generated and human-authored content, focusing on whether LLMs exhibit higher or lower emotional intensity. Emotional content was quantified using a probabilistic model assigning normalized scores (0 to 1) to each emotional category, interpreted as intensity scores \cite{miyazaki2024impact}. These scores were grouped into five intensity levels. Statistical analysis included ANOVA to compare group differences and Tukey’s post-hoc test for significant pairwise variations\cite{article}.

\begin{table}[]
\caption{ANOVA test of different model emotions}
\begin{tabular}{cccc}
Platform                 & Emotions & F statistic                  & P value                               \\ \hline
\multirow{7}{*}{Reddit}  & anger    & 114.2381                     & \textless{}0.001                     \\
                         & joy      & 80.899                       & \textless{}0.001                     \\
                         & disgust  & 29.8564                      & \textless{}0.001                     \\
                         & surprise & 24.9967                      & \textless{}0.001                     \\
                         & fear     & 17.3482                      & \textless{}0.001                     \\
                         & sadness  & 5.5149                       & \textless{}0.001                     \\
                         & neutral  & \multicolumn{1}{l}{325.2502} & \multicolumn{1}{l}{\textless{}0.001} \\ \hline
\multirow{7}{*}{Twitter} & anger    & 202.1090                     & \textless{}0.001                     \\
                         & joy      & 34.4011                      & \textless{}0.001                     \\
                         & disgust  & 22.4241                      & \textless{}0.001                     \\
                         & surprise & 51.5744                      & \textless{}0.001                     \\
                         & fear     & 75.0905                      & \textless{}0.001                     \\
                         & sadness  & 9.2834                       & \textless{}0.001                     \\
                         & neutral  & 552.6669                     & \textless{}0.001                    

\label{ANOVA Test}
\end{tabular}
\end{table}

Analysis of variance (ANOVA) results presented in Table \ref{ANOVA Test} indicate statistically significant differences ($P < 0.01$) across the nine groups in emotional intensity values for anger, disgust, fear, sadness, joy, surprise and neutral on Twitter. Similar significant variations were also observed in the Reddit dataset for the seven emotions mentioned above. Tukey's post-hoc analysis identified several significant differences across three comparison categories: within-model, between-model, and model-to-human comparisons.


Each emotion score represents its degree of emotion, and the higher the score, the stronger the emotion. As shown in the table\ref{comparison_models}, intra-model comparisons indicate that emotional scores for certain emotions—such as anger and fear—are generally higher in continuation tasks than in response tasks. This suggests that LLMs tend to adopt a more neutral emotional tone when operating as conversational agents. Furthermore, when compared to the original human-authored texts,shown in the table\ref{comparison2human}, most LLM-generated outputs exhibit lower emotional intensity, particularly in dimensions such as anger, fear, disgust, and joy. This may indicate that LLMs express emotions in a more restrained and subdued manner relative to humans.

Cross-model comparisons also reveal that emotional expression varies significantly depending on task type and data source. For instance, within the Llama model family, Llama 3 demonstrates higher emotional scores than Llama 3.3, suggesting a greater tendency toward heightened emotional expression.

\begin{sidewaystable}[htbp]
\caption{Tukey's post-hoc test of LLM-generated content emotion values}
\label{comparison_models}
\centering
\begin{threeparttable}
\resizebox{\textwidth}{!}{%
\begin{tabular}{l p{1.4cm} p{1.4cm} p{1.4cm} p{1.6cm} p{0.1cm} p{1.6cm} p{1.6cm} p{1.6cm} p{1.6cm} p{1.6cm} p{1.8cm} p{1.8cm} l}
\multicolumn{1}{c}{\textbf{Emotion}} & \multicolumn{4}{c}{\textbf{Within Group}\tnote{1}} & \multicolumn{1}{c}{} & \multicolumn{7}{c}{\textbf{Between Groups}\tnote{2}} & \multicolumn{1}{c}{\textbf{Platform}} \\ \hline
                                     & Gemma Con vs Resp & Llama Con vs Resp & Claude Con vs Resp & Llama3.3 Con vs Resp & Gemma Con vs Llama Con & Gemma Con vs Claude Con & Llama Con vs Claude Con & Llama Con vs Llama3.3 Con & Gemma Resp vs Llama Resp & Gemma Resp vs Claude Resp & Llama Resp vs Claude Resp & Llama Resp vs Llama3.3 Resp & \\
\cline{2-13}
anger                                & \textgreater{}*** & \textless{}***    & \textgreater{}***  & -                    & \textgreater{}**       & -                       & \textgreater{}***       & -                         & \textless{}***           & \textless{}*              & \textless{}***            & \textgreater{}***           & \multirow{7}{*}{Reddit} \\
fear                                 & \textless{}***    & -                 & \textless{}**      & \textless{}***       & \textless{}***         & \textgreater{}***       & -                       & \textgreater{}*           & -                        & -                         & -                         & -                           & \\
sadness                              & \textgreater{}**  & \textless{}***    & \textless{}**      & -                    & \textgreater{}***      & -                       & \textgreater{}***       & -                         & \textless{}**            & \textgreater{}***         & \textgreater{}***         & -                           & \\
joy                                  & -                 & -                 & \textgreater{}***  & \textgreater{}***    & \textgreater{}***      & -                       & \textgreater{}***       & -                         & \textgreater{}*          & \textless{}**             & -                         & \textgreater{}***           & \\
surprise                             & -                 & \textgreater{}**  & -                  & -                    & \textgreater{}***      & \textless{}**           & -                       & -                         & \textgreater{}***        & \textless{}*              & -                         & -                           & \\
disgust                              & \textless{}*      & \textless{}***    & \textless{}**      & \textless{}***       & -                      & -                       & -                       & \textgreater{}***         & -                        & -                         & -                         & -                           & \\
neutral                              & \textgreater{}*** & \textgreater{}*** & \textgreater{}**   & \textgreater{}***    & -                      & \textless{}***          & \textless{}***          & \textgreater{}***         & \textless{}***           & \textgreater{}*           & \textless{}***            & \textgreater{}***           & \\ \hline
anger                                & -                 & -                 & \textgreater{}***  & \textgreater{}***    & \textgreater{}***      & -                       & \textless{}***          & -                         & \textgreater{}*          & \textgreater{}**          & -                         & \textgreater{}***           & \multirow{7}{*}{Twitter} \\
fear                                 & \textgreater{}*** & \textgreater{}*** & \textgreater{}**   & \textgreater{}***    & -                      & \textgreater{}***       & \textgreater{}***       & \textgreater{}***         & \textless{}***           & \textless{}***            & \textgreater{}***         & \textgreater{}***           & \\
sadness                              & \textless{}***    & -                 & \textless{}*       & \textless{}***       & \textless{}***         & \textless{}***          & -                       & \textgreater{}*           & -                        & -                         & -                         & -                           & \\
joy                                  & -                 & \textgreater{}*** & -                  & -                    & \textgreater{}***      & \textgreater{}**        & -                       & -                         & \textgreater{}***        & \textgreater{}*           & -                         & -                           & \\
surprise                             & \textgreater{}*   & \textless{}***    & \textless{}**      & -                    & \textgreater{}***      & -                       & \textless{}***          & -                         & \textless{}**            & \textless{}***            & \textless{}***            & -                           & \\
disgust                              & \textgreater{}*** & \textless{}***    & \textgreater{}***  & -                    & \textgreater{}**       & -                       & \textless{}***          & -                         & \textless{}***           & \textgreater{}*           & \textgreater{}***         & \textgreater{}***           & \\
neutral                              & \textless{}*      & \textless{}***    & \textless{}**      & \textless{}***       & -                      & -                       & -                       & \textgreater{}***         & -                        & -                         & -                         & -                           & \\
\end{tabular}
} 
\begin{tablenotes}
\footnotesize
\item[1] The comparison of the emotion values of continuation and response content generated by the same model.
\item[2] The comparison of the emotion values of the same task but generated by different model.
\item[3] The stars indicate the \textit{p} values of the Mann–Whitney U test: *** for $p<0.001$, ** for $p<0.01$, and * for $p<0.05$.
\item[4] The symbolic $<$ and $>$ indicate that the data in the previous column is less than or greater than the data in the next column. The “-” means there is no statistically \\significant difference between the two columns of data.
\end{tablenotes}
\end{threeparttable}
\end{sidewaystable}

\begin{sidewaystable}[htp]
\caption{Tukey's post-hoc test of LLM-generated content and human text emotion values}
\label{comparison2human}
\centering
\begin{threeparttable}
\resizebox{\textwidth}{!}{%
\begin{tabular}{cccccccccc}
\textbf{Platform}        & \textbf{Emotions} & \textbf{Gemma Con vs Original\tnote{1}} & \textbf{Gemma Resp vs Original} & \textbf{Llama Con vs. Original} & \textbf{Llama Resp vs. Original} & \textbf{Claude Con vs Original} & \textbf{Claude Resp vs Original} & \textbf{Llama3.3 Con vs Original} & \textbf{Llama3.3 Resp vs Original} \\ \hline

\multirow{7}{*}{Reddit}  & anger             & \textless{}***                 & \textless{}***                  & \textless{}***                  & \textless{}***                   & \textless{}***                  & \textless{}***                   & \textless{}***                    & \textless{}***                     \\
                         & fear              & \textless{}***                 & -                               & -                               & -                                & -                               & -                                & \textless{}***                    & -                                   \\
                         & sadness           & -                              & -                               & -                               & -                                & \textgreater{}*                 & \textgreater{}***                & -                                 & -                                   \\
                         & joy               & -                              & -                               & \textless{}***                  & -                                & -                               & \textless{}**                    & \textless{}**                     & \textless{}***                     \\
                         & surprise          & -                              & -                               & \textless{}***                  & \textless{}***                   & -                               & -                                & \textless{}***                    & -                                   \\
                         & disgust           & \textgreater{}***              & \textgreater{}***               & \textgreater{}***               & \textgreater{}***                & \textgreater{}**                & \textgreater{}***                & -                                 & \textgreater{}***                  \\
                         & neutral           & \textgreater{}***              & \textless{}***                  & \textgreater{}***               & \textless{}**                    & \textless{}***                  & \textless{}***                   & -                                 & \textless{}***                     \\ \hline

\multirow{7}{*}{Twitter} & anger             & -                              & -                               & \textless{}***                  & -                                & -                               & \textless{}**                    & \textless{}**                     & \textless{}***                     \\
                         & fear              & \textgreater{}***              & \textless{}***                  & \textgreater{}***               & \textless{}**                    & \textless{}***                  & \textless{}***                   & -                                 & \textless{}***                     \\
                         & sadness           & \textless{}***                 & -                               & -                               & -                                & -                               & -                                & \textless{}***                    & -                                   \\
                         & joy               & -                              & -                               & \textless{}***                  & \textless{}***                   & -                               & -                                & \textless{}***                    & -                                   \\
                         & surprise          & -                              & -                               & -                               & -                                & \textgreater{}*                 & \textgreater{}***                & -                                 & -                                   \\
                         & disgust           & \textless{}***                 & \textless{}***                  & \textless{}***                  & \textless{}***                   & \textless{}***                  & \textless{}***                   & \textless{}***                    & \textless{}***                     \\
                         & neutral           & \textgreater{}***              & \textgreater{}***               & \textgreater{}***               & \textgreater{}***                & \textgreater{}**                & \textgreater{}***                & -                                 & \textgreater{}***                  
\end{tabular}
}
\begin{tablenotes}
\footnotesize
\item[1] The comparison of the emotion values of the LLM-generated content by different model and human text.
\item[2] The stars indicate the \textit{p} values of the Tukey's post-hoc test: *** for $p<0.001$, ** for $p<0.01$, and * for $p<0.05$.
\item[4] The symbolic ``$<$ and $>$'' indicate that the data in the previous list is less than or greater than the data in the next list. The ``-'' means there is no statistically \\significant difference between the two columns of data.
\end{tablenotes}
\end{threeparttable}
\end{sidewaystable}



These findings suggest two key insights: first, LLMs demonstrate systematic suppression of certain negative emotions, particularly in continuation tasks; second, the response task appears to operate under distinct generative mechanisms, resulting in differential emotional expression patterns. Furthermore, the consistent reduction in optimism across all LLM-generated texts relative to human-authored content indicates a systematic constraint in LLMs' capability to fully capture and convey positive emotional states.


\subsection{Evaluating Semantic Consistency of LLM-Generated Content in Social Media Contexts}

In LLM-human interactions, we analyzed models' ability to maintain topical coherence and contextual relevance using scores labeled by LLM-as-judge~\cite{zheng2023judging} to measure semantic and logic alignment with the original content, providing a framework for evaluating semantic and logic fidelity across contexts.


\begin{figure}[htb]
    \centering
    \includegraphics[width=\linewidth]{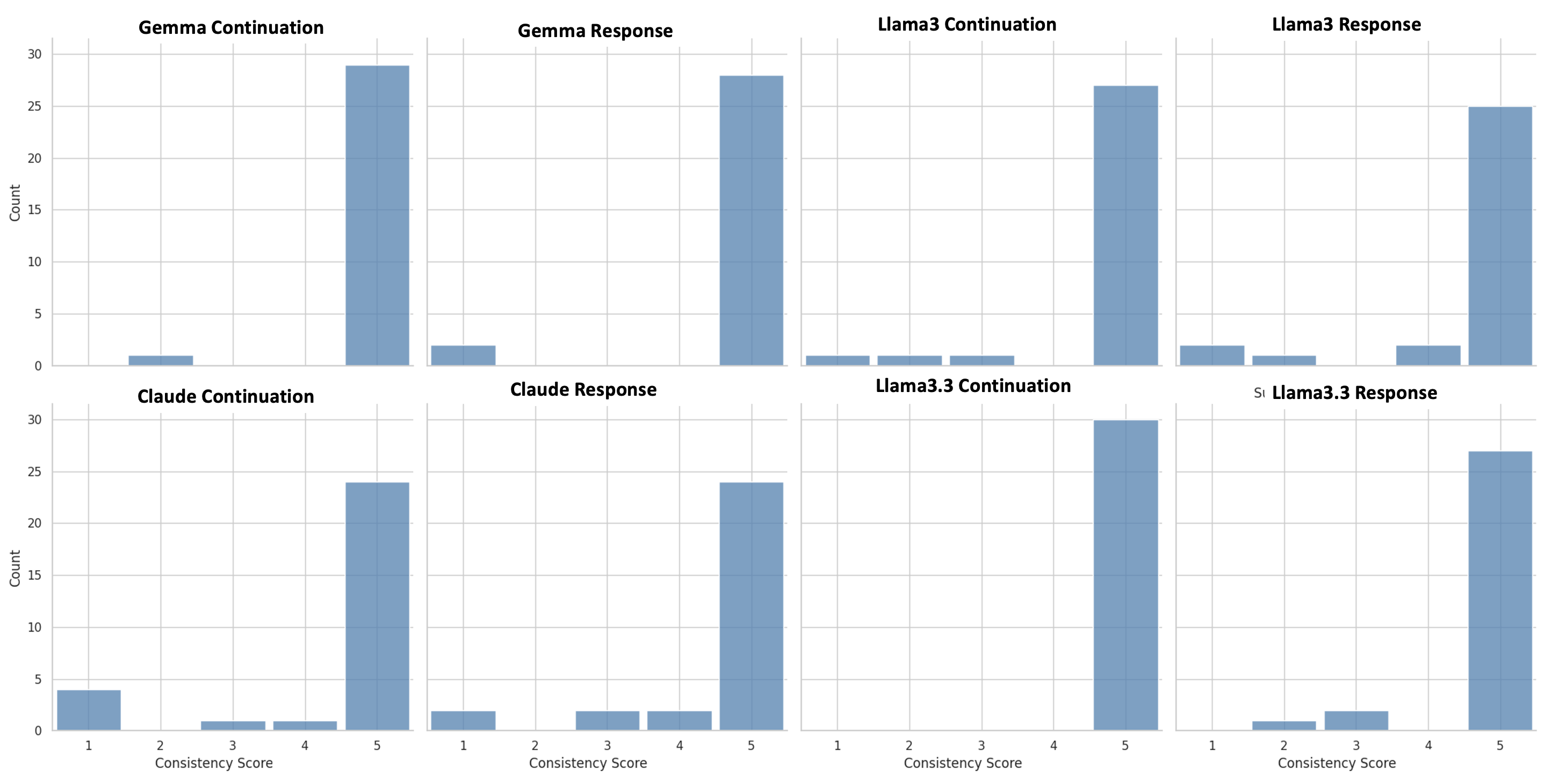}
    \centerline{\textbf{a}}

    \vspace{0.5em}  

    \includegraphics[width=\linewidth]{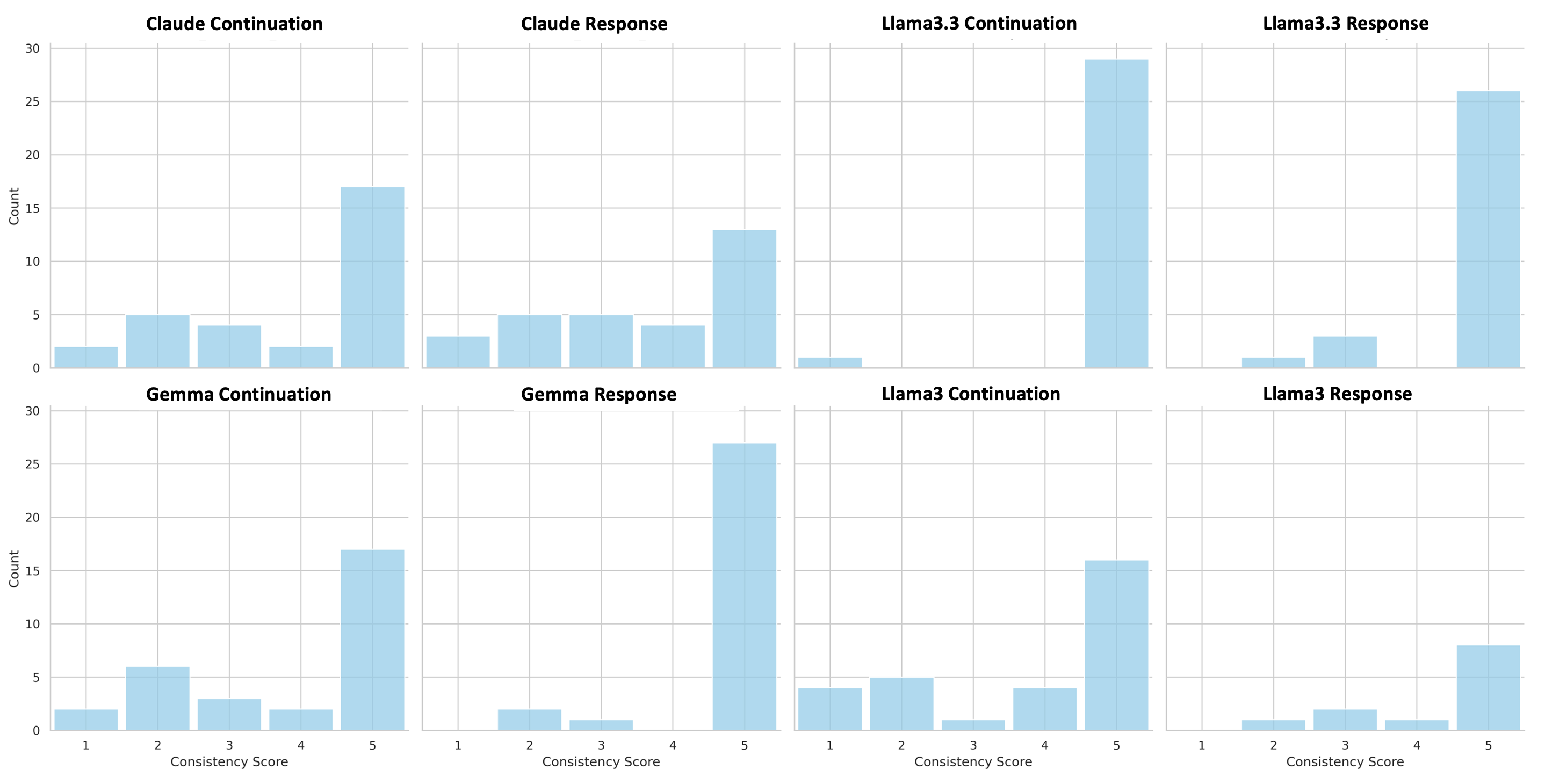}
    \centerline{\textbf{b}}

    \caption{Distribution of faithfulness and consistency scores for LLM-generated content, visualized using stacked histograms comparing Gemma, Llama, and Claude models' continuations and responses across Reddit (panel \textbf{a}) and Twitter (panel \textbf{b}) platforms. The x-axis denotes rating categories, with 5 representing 'Excellent', 4 'Good', 3 'Fair', 2 'Poor', and 1 'Very Poor'.}
    \label{fig:faithfulness_consistency}
\end{figure}


Figure~\ref{fig:faithfulness_consistency} plotted the frequency of each score (ranging from 1 to 5) separately for each task. This allows us to examine whether certain tasks tend to produce more consistent model outputs. The results suggest that while most tasks concentrate around higher scores, there are subtle differences in how each subtask is rated by the evaluator. We find that most models are capable of generating high-quality content in both response and continuation tasks, indicating that current LLMs possess strong capabilities in understanding and interacte with human language. However, subtle differences remain across models. For instance, the Llama 3.3 model demonstrates consistently strong performance in content generation across both types of datasets. Comparized with Llama 3' result, this may also indicate to some extent the impact of different model sizes on the quality of generated content. Additionally, model performance varies across platforms. For example, the Claude model shows higher fidelity and consistency in its outputs on the Reddit dataset, which may be attributed to the more complex linguistic environment of Twitter.

\section{Discussion}

This study utilized the Twitter and Reddit datasets on climate change to systematically evaluate three aspects of LLM-generated text versus original social media posts: (1) emotional consistency in both continuation and response tasks (and differences in different models), (2) emotional intensity across those tasks, and (3) logic similarity and context-awareness between model outputs and original social media texts. By combining these analyses, we aim to uncover the potential effects on emotional dynamics that LLMs have when implemented in online discussion.

In the continuation task, Gemma consistently transformed most emotions toward anger, showing a bias toward negative expression and intensifying anger. However, it maintained optimism and surprise, suggesting some ability to preserve emotional valence. On the other hand, Llama better preserved original emotions like anger, anticipation, fear, optimism, and sadness, with fewer emotional shifts, reflecting stronger emotional continuity (RQ1).



By combining the comparative analyses of emotional consistency and intensity, this study demonstrates that in both response and continuation tasks, LLMs systematically shift emotional content toward neutrality and lower its intensity across most emotional categories compared to original texts. Particularly, generated outputs display a higher prevalence and intensity of ``neutral" manner, especially in response tasks rather than continuation tasks. This pattern shows that LLMs do have effect on the change of emotion when engaged in online discussion, and they're turning the discussion in a emotionally-moderate way.


As aforementioned, emotion plays a critical role in navigating social media discussion, especially in controversial topics like climate change. While the climate change discussion online is quite polarized \cite{treen2022discussion}, such a tendency to neutrality could help defuse heated exchanges and foster calmer, less contentious dialogue. Hence, even though LLMs cannot keep the emotional consistency as human, its preferences  toward moderation could benefit the highly polarized and emotionally debatable topics online, and lead the discussion more neutrally.

However, beyond this ``neutral" tendency, what should be aware is that the negative emotions like ``anger" and ``fear" also showed increase in LLM-generated continuations and responses in most of the cases, which indicates that these models can amplify such emotional cues. Anger and fear are among the most salient emotional frames in climate discourse \cite{nabi2018framing, davidson2022emotional}, and are significantly associated with climate activism, engagement, and has great effect on people’s attitudes and behaviors \cite{stanley2021anger, gregersen2023strength}. An uptick in fear and anger in LLM outputs may therefore shape users’ perceptions of climate change. Considering the echo chamber effect on social media, LLMs' propensity of fear and anger could increase people's perceived threat by climate change, which in turn leads to higher awareness and action intentions towards climate crisis.

In terms of semantic consistency, the results of LLM-as-judge (Figure 6) show that LLMs generated text could generally keep the consistency to social media texts. This comparative findings proves a critical evidence from the perspective of emotion that LLMs could generally “understand” the  context of a social media post and produce continuations or replies that align both thematically and logically with the source content. However, the consistency varies due to different types of LLMs, while most of the scores are high, some specific model like Claude's performances are dispersed, featuring a nontrivial tail into other scores. It indicates that some models may be more sensitive to prompt phrasing or exhibit less robust world-knowledge integration. The findings from semantic consistency imply that LLMs nowadays can generally preserve semantic and logic coherence in social-media dialogue, which makes the LLMs-generated content hard to distinguish on social media.

Taken together, the emotional and semantic consistency examined in this study could be seen as a technological improvement of context-awareness in LLMs' text generation capabilities. However, it could also raise practical concerns to general social media users for the risks of LLMs being manipulated for some malicious or intentional purposes, such as misinformation.

Effectively managing public emotions during controversial discussion online is crucial for governance of public opinion. Recent advances in Artificial Intelligence Generated Content (AIGC), driven by Generative AI (GAI) technology, have garnered attention beyond computer science \cite{cao2023comprehensivesurveyaigeneratedcontent}. Given the increasing integration of LLMs into daily life, their emotional characteristics significantly influence opinion leadership, as emotional content shapes public perception and discourse framing.

Future implications suggest that while LLMs generate semantically coherent content, there is potential to improve alignment with nuanced human contexts. Research should focus on refining LLMs' understanding of implicit meaning and contextual subtleties to enhance user experience and broaden application domains.

\subsection{Limitation and Future Work}
This study enhances understanding of emotional dynamics in human-AI interactions while highlighting opportunities for future research to address current limitations.

First, regarding experimental design, our analysis relied on Reddit and Twitter data. Platforms like YouTube, Instagram, and TikTok exhibit unique user behaviors and content structures \cite{Voorveld02012018}. 

Second, the explaination for the findings in emotional inconsistency of LLMs requires further exploration. While we revealed that emotions vary both across different LLMs and across generation tasks, for example, the ``neutral" emotion take the major role among all emotions in all cases but differs between continuation and response tasks and LLMs, likewise, the overall proportion of ``anger” significantly increases in all cases, and different models exhibit distinct emotional profiles. However, because the training data and internal weights of these models are not publicly available, we cannot at present pinpoint the underlying causes of these differences—whether they stem from pretraining data biases,  fine-tuning strategies, or discussion context. Future work should work on the interpretability analyses, controlled comparisons, or replication on open datasets, to investigate the mechanisms driving these emotion changes.

Third, this study may have potential data contamination due to the use of publicly available data from Twitter and Reddit. Twitter and Reddit data are publicly accessible and may overlap with the LLMs’ pre-training corpora, model outputs could partly reflect memorized patterns rather than genuine reasoning in real-world contexts, which limits the generalization of the results. While we cannot fully quantify the contamination, we acknowledge its impact and encourage future research to explore methods for mitigation—such as by using post-training data that is demonstrably out-of-scope of LLM training sets, or by focusing on data published after known pretraining cutoff dates. Besides, social media-derived content may include automated posts and social bot interactions, potentially skewing results. Future research should adopt stricter data validation protocols to ensure datasets quality and reduce confounding variables.

Fourth, since human expert evaluation is expensive, we adopt a hybrid evaluation strategy that combines human judgment with LLM assistance based on a small but diverse sample across platforms, tasks, and models. While this approach achieves high-quality and interpretable scores, we acknowledge that the limited sample size may not capture the full distribution of consensus behaviors across the datasets. Future work could extend this approach, such as scalable pure LLM evaluation and automatic divergence detection for selective human review.

\section{Conclusion}



In this study, we evaluated how large language models handle emotional and semantic content in social-media contexts. By comparing four representative models—Gemma, Llama3, Claude, Llama3.3—on both continuation and response tasks, we uncovered distinct patterns in their emotional behaviors and contextual understanding.

Across both continuation and response tasks, LLMs tended to shift emotional content toward neutrality and generally reduced emotional intensity. This effect was more pronounced in response tasks, suggesting that LLMs favor emotionally moderate language when addressing existing posts. LLM-as-judge evaluations revealed that generated texts largely preserved thematic and logical coherence with source texts. While most models scored consistently high, variability in some (e.g., Claude) highlights sensitivity to prompt phrasing and model-specific robustness.
The neutrality bias of LLMs could help defuse polarization and foster calmer dialogue around contentious topics. Conversely, the amplification of anger and fear poses both opportunities such as mobilization and risks like manipulation or echo-chamber reinforcement.

These findings reveal key emotional dynamics between LLMs, offering insights for improving their design and use in emotion-sensitive applications.




\bibliography{main}

\end{document}